# A Syllable-Structured, Contextually-Based Conditionally Generation of Chinese Lyrics

Xu Lu[1,2], Jie Wang[2], Bojin Zhuang[2], Shaojun Wang[2] and Jing Xiao[2]

[1] Beijing University of Posts and Telecommnications, China
[2] Ping An Technology (Shenzhen) Co., Ltd, China
lx0523@bupt.edu.cn, photonicsjay@163.com,
zhuangbojin232@pingan.com.cn, wangshaojun851@pingan.com.cn,
xiaojing661@pingan.com.cn

**Abstract.** This paper presents a novel, syllable-structured Chinese lyrics generation model given a piece of original melody. Most previously reported lyrics generation models fail to include the relationship between lyrics and melody. In this work, we propose to interpret lyrics-melody alignments as syllable structural information and use a multi-channel sequence-to-sequence model with considering both phrasal structures and semantics. Two different RNN encoders are applied, one of which is for encoding syllable structures while the other for semantic encoding with contextual sentences or input keywords. Moreover, a large Chinese lyrics corpus for model training is leveraged. With automatic and human evaluations, results demonstrate the effectiveness of our proposed lyrics generation model. To the best of our knowledge, there is few previous reports on lyrics generation considering both music and linguistic perspectives.

## 1    Introduction

Natural language generation (NLG) plays an important role in machine translation, dialogue generation and other fields. In the recent years, owing to the fast rise of deep learning, RNN and other alternate neural networks are often used in NLG applications. In particular, an encoder-decoder based sequence-to-sequence (Seq2Seq) generation framework has been widely applied in various NLG problems including poetry and short essays generation. The main idea behind the Seq2Seq model is to encode input sequences into a fixed-length dense vector, and then decode corresponding sequences from this contextual vector. Moreover, attention mechanism has also been incorporated into this architecture to learn to soft alignments between contextual semantics.

Given a piece of melody, automatic lyrics generation is a challenging task. Completely different from prose text, lyrics generation should include both knowledge and consideration of music-specific properties including melody structure, rhythms, etc. For instance, word boundaries in lyrics and the rests in a melody should be consistent. As depicted in **Fig. 1**, it sounds unnatural if a single syllable spans beyond a long melody rest. During the procedure of lyrics writing, such constraints in content and lexical selection could impose extra cognitive loads.

Because of Chinese language specificity, one Chinese character represents one syllable. Therefore, beat patterns of melody can be interpreted as character number of lyrics and its fine sections. Different from regular poems, the form of lyrics is more

*(You ask me how deep I love you, how much I love you.)*

**Fig. 1**. Structural alignment between lyrics and melody of a Chinese popular song.

free. To address this issue, this paper proposes a novel two-channel Seq2Seq for lyrics generation, which combines both syllable-pattern and contextual semantic information. With attention mechanism, singable lyrics can be generated and perfectly matched with the original melody.

The remainder of this paper is organized as follows. In Section 2, background about NLG is introduced. In Section 3, our lyrics generation model is described at detail. Section 4 discusses the model structure and experimental results. Section 5 concludes the work. Main contributions of our work are also listed:

- We propose a syllable-structured lyrics generation model, considering both music specialty and language attribute simultaneously with a two-channel encoder.
- To improve the singability of generated lyrics, the beat pattern of melody has been approximately interpreted as syllable structural information.
- To enhance the coherence and entirety of generated lyrics, the contextually-based conditional generation model can take in previous sentences or keywords.
- We leverage a large Chinese lyrics corpus of 300,000 pop songs to pre-train this model.

## 2    Background

### 2.1    Prior Work

Automatic text generation has been always a popular but challenging research topic. Recent work (Oliveira 2012) has been conducted to address this problem with grammatical and semantic templates. Statistical machine translation methods (He 2012) have also been exploited, in which each new line is considered as a "translation" of the previous line. Deep learning has also been proposed for language generation. For instance, an attention-based bidirectional RNN model (Yi 2016) was proposed for generating 4-line Chinese poems. Except for Chinese regular poems, Chinese iambic poems with free forms has also been demonstrated (Wang 2016). Moreover, the language model of LSTM (Potash 2015) was used to generate rap lyrics with a desired style, but failed to control the structure flexibly. In order to consider music properties, a melody-conditioned language model (Watanabe 2018) was proposed to generate Japanese lyrics. However, lyric-melody aligned data was really rare and highly cost if labelled by experts. Moreover, a RNN based language model is really difficult to capture long-term contextual information and hard to generate coherent multi-paragraph lyrics. To address this issue, we propose a two-channel Seq2Seq model which can contextually generate texts by taking in previous sentences or keywords.

### 2.2 RNN Encoder-Decoder

The RNN encoder-decoder framework (Sutskever 2014) is firstly introduced, of which the encoder and decoder are two separate RNN modules. The encoder converts a sequence of input ($x_1$, ..., $x_t$) to a contextual dense vector $c$. Vector $c$ encodes information of the whole source sequence, and is incorporated into decoder to generate the target output sequence. Thus, the probability distribution of prediction is defined as:

$$P(Y) = \prod_{t=1}^{T} P(y_t|y_{t-1}, c) \tag{1}$$

where $y_{t-1}$ represents the generated output sequence prior to time step $t$. Different from an RNN based language model, the encoder-decoder model is capable of mapping sequence to sequence even from different domains. To apply explicit alignment between source and target sequences, attention mechanism is incorporated into this model.

## 3   Proposed Methods

In this section, a baseline model of lyric generation with an attention based encoder-decoder architecture is described. Following that, we describe the proposed method to control the generation of syllable structure and content with a multi-channel Seq2Seq model.

### 3.1   Baseline model

In the encoder of the baseline model, a bidirectional RNN. is used, which has been successfully applied in text generation and spoken language understanding. In addition, LSTM is used as the basic RNN unit because of its better long-term dependencies than vanilla RNN.

During the lyrics generation, context-aware generation is realized by inputting ($x_1$, ..., $x_t$). The bidirectional LSTM reads the source word sequence forward and backward. The forward RNN reads the word sequence in its original order and generates a hidden state $h_{fi}$ at each time step. Similarly, the backward RNN reads the word sequence in the reverse order and generate a sequence of hidden states ($h_{bT}$, ..., $h_{b1}$). The final encoder hidden state $h_i$ at each time step i is a concatenation of the forward state $h_{fi}$ and backward state $h_{bi}$ i.e. $h_i = [h_{fi}, h_{bi}]$.

Therefore, last state of the forward and backward RNN carries information of the entire source sequence. We use the last state of the backward encoder as the initial decoder hidden state following the approach (Bahdanau 2014). The decoder is a unidirectional LSTM. At each decoding step i, the decoder state $s_i$ is calculated as a function of the previous decoder state $s_{i-1}$, the previous predicted token $y_{i-1}$, the encoder hidden state $h_i$ and the context vector $c_i$:

$$s_i = f(s_{i-1}, y_{i-1}, h_i, c_i) \qquad (2)$$

where the context vector $c_i$ is computed as a weighted sum of the encoder states $h = (h_1, ..., h_T)$ [11]:

$$c_i = \sum_{j=1}^{T} \alpha_{i,j} h_j \qquad (3)$$

And
$$\alpha_{i,j} = \frac{\exp(e_{i,j})}{\sum_{k=1}^{T} \exp(e_{i,k})}, \qquad e_{i,k} = g(s_{i-1}, h_k) \qquad (4)$$

where $g$ is a feed-forward neural network. At each decoding step, the explicit aligned input is the encoder state $h_i$. The context vector $c_i$ provides extra information to the decoder and can be seen as a continuous bag of weighted features ($h_1, ..., h_T$).

### 3.2 Multi-Channel Seq2Seq with Attention

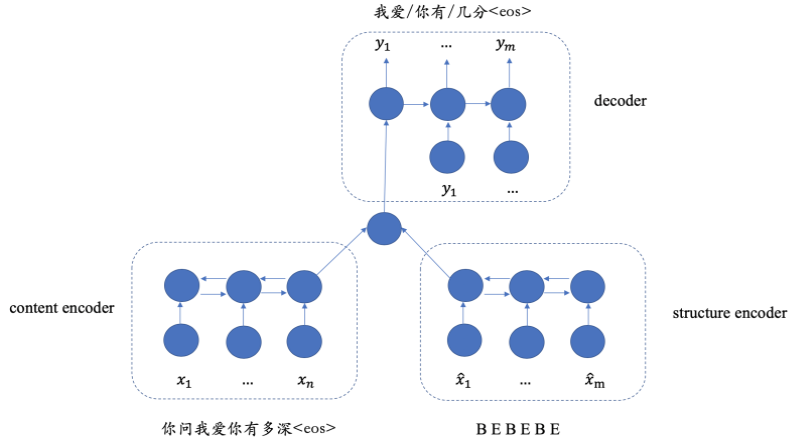

**Fig. 2.** Scheme of the multi-channel Seq2Seq generation model with structure and content encoders. Tokens of 'S', 'B', 'M', 'E' represent the syllable structure of lyrics, meaning the start of a sentence, the beginning, middle and end of a music segmentation of a melody.

As shown in **Fig. 1**, there is a finer structural alignment between lyrics and melody, except for sentence length. To be consistent with the beat pattern of melody, finer control of lyrics structures is required. Due to specialty of Chinese, phrase groups with certain character numbers retain both semantic and beat pattern information. To fuse both music and language into sentence decoding, we propose a multi-channel Seq2Seq model as shown in **Fig. 2**. Two different Bi-LSTMs read the structural token sequence and previous sentences, while text generation is only modeled with one forward LSTM. When the attention mechanism is enabled, the context vector $c_i$ provides partial information from input sequences that is used together with the aligned hidden state $h_i$ for generating Chinese characters. Different from the contextual vector in Section 3.1, $c_i$ in this model is calculated based on the concatenation of the two encoders' hidden states. But the initial state of the decoder is still same with the last state of content encoder.

Other than syllable structural encoding, two approaches of content encoding have been conducted to determine new sentence generation. In one case, two neighboring sentences extracted from lyrics are processed as previous and next sequence pairs. Similar to a kind of monolingual translation, next sentence can be generated by taking the previous one as input. In the other case, one keyword was retrieved from one lyric sentence to form training corpus of keyword-sentence pair. For keyword retrieval, a text-rank algorithm was used.

## 4    Experiments

### 4.1    Data

A large corpus of Chinese lyrics of 160,000 songs has been prepared to pre-train our lyrics generation language model. From this corpus, corresponding music notations of 50,000 songs have been manually interpreted with crowdsourcing. Among them, 4.15 million previous-next sentence/keyword-sentence pairs have been accumulated for model training. 10,000 lyric sentences are used for evaluation of model.

### 4.2    Training Procedure

LSTM cell is used as the basic RNN unit in all models, of which the dimension size of hidden state is 128. And then 4 layers of LSTM networks are used in the proposed models. Embedding size of 128 of Chinese characters are randomly initialized and then fine-tuned by training with mini-batch size of 16. Dropout rate of 0.3 is used to the non-recurrent connections for model regularization. Maximum norm for gradient clipping is set to 1. Adam method is used for model optimization and Bahdanau attention is applied. Schedule sampling with probability of 0.1 instead of teaching force is used for preparing ground truth. For inference, beam search decoding with beam width size of 35 is used.

### 4.3    Evaluation Metric

**Automatic Evaluation**

To evaluate our proposed model, a test lyric corpus has been selected as ground truth reference. For the melody structure control, the melody-alignment accuracy between prediction and ground truth will be calculated. In the case of semantic prediction, BLEU scores of generations will be computed. Even if BLEU is not a suitable metric for this NLG task, we believe that it can still reflect the semantic coherence and relevance of generation. Note that Bi-gram is the max length of n-gram. And BLEU can reflect the degree of control of the content which is very important.

**Human Evaluation**

Since lyrics generation belongs to literature creation, human evaluation might be a better way for performance evaluation. Following the reference (Yi 2016), three criteria has been designed: Fluency (fluency of generated sentences), Meaningfulness (do generated lyrics convey some certain messages and the contextual relevance?), Diversity (do generated sentences often show similar phrases or word?), and Entirety (general impression on sentences). Five thousand of generated lyrics samples were cross-scored by five Chinese language experts with score range of 0 to 5.

### 4.4 Results and Analysis

**Automatic Evaluation Result**

Compared to previous work, main contribution of this paper is to fuse syllable-structural control into text generation. As seen in Table 1 and Fig. 3, the multi-channel generation model can completely control the output sentence length, and melody-matching accuracy is also very high. With a beam-search decoder, the baseline model tends to generate sentences with shorter length and results in lower BLEU scores. However, BLEU of the keyword-aware generation model (KG) is higher than others including the sentence-aware generation model (SG), because the encoded keyword provides more context information, which offers a promising approach for structure and semantic control in NLG.

**Table 1**. Accuracy of Automatic Evaluation

| Models | Length Control | Melody Matching | BLEU |
|---|---|---|---|
| Baseline | 8.20% | 2.00% | 2.83% |
| SG | **100.00%** | **87.60%** | 4.61% |
| KG | 99.93% | 83.85 | **16.22%** |

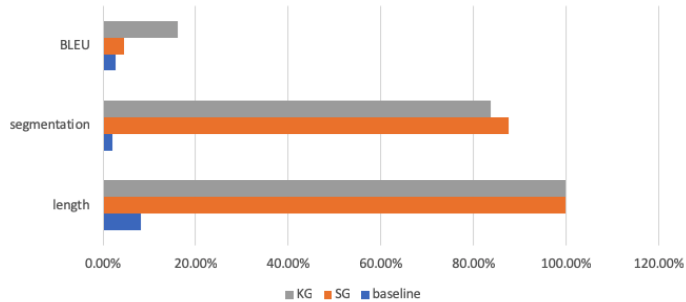

**Fig 3.** Overall comparison of automatic evaluation.

### 4.5 Human Evaluation Result

As seen from **Table 2** and **Fig 4**, our model performs much better than the baseline. One main reason is that the generic Seq2Seq baseline model tends to generate short sentences, often less than five words, which decreases the meaningfulness and increases repeatability of generated sentences. While with our generation model, two encoders of structure and content will mutually promote the effect of decoding referred to (Ghazvininejad 2016).

**Table 2.** Score of human evaluation

| Models | Fluency | Meaningful | Diversity | Entirety |
|---|---|---|---|---|
| Baseline | 3.01 | 2.11 | 1.2 | 2.11 |
| SG | 4.21 | 3.74 | 3.92 | 4 |
| KG | **4.25** | **4.54** | **4.43** | **4.4** |

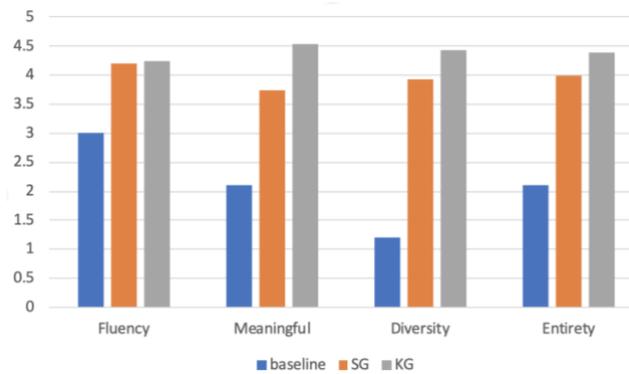

**Fig. 4.** Human evaluation of three models.

Moreover, the KG model owns higher scores than SG, because the encoded keyword represents global context of next generation, resulting in the increase of meaningfulness and diversity scores. Overall, both the KG and SG models have achieved satisfactory results. Furthermore, during the generation of an entire lyrics, it is too cumbersome to give one keyword for each sentence. Therefore, in the real lyrics generation, SG and KG models fuse together to compose a full-paper lyrics samples. **Figure 5** shows a typical example.

**Fig. 5.** Generated lyrics with tune of Chinese song "Water for Forgotten Love" (忘情水).

## 5 Conclusion and Future Work

In this paper, we consider lyrics generation as a sequence-to-sequence learning problem, and propose a two-channel Seq2Seq generation model conditioned on input melody. Better than the baseline, our model jointly learns contextual information and syllable structures, verified by both automatic and human evaluation. Moreover, our proposed approach can be extended to other kinds of language structure control, including poetry rhythm and specific language templates.

To generate perfect lyrics, there are still further works for model polishing. The topic and emotion of the entire lyrics is difficult to handle only through encoding keywords and contextual sentences. Therefore, encoding channels can be scaled to fuse more controlling information, including global theme, sentiment or other literature styles.

## 6 Acknowledgement

This work was supported by Ping An Technology (Shenzhen) Co., Ltd, China.